\documentclass{article}
\usepackage{spconf,amsmath,graphicx}

\usepackage{booktabs}
\usepackage{graphicx}
\usepackage{array}
\usepackage{multirow}
\usepackage{tabularx}
\usepackage{amssymb}
\usepackage{xcolor}
\definecolor{mygreen}{RGB}{0,100,0}

\newcommand{\smallgreen}[1]{\textcolor{mygreen}{\scalebox{0.8}{#1}}}

\title{Improving Cross-domain Few-shot Classification with Multilayer Perceptron}

\name{\parbox{\linewidth}{\centering 
Shuanghao Bai$^{1}$, 
Wanqi Zhou$^{1}$, 
Zhirong Luan$^{2}$, 
Donglin Wang$^{3,^*}$, 
Badong Chen$^{1,^*}\thanks{*Corresponding author.}$
}
\thanks{$^{\dag}$This work was supported by the National Natural Science Foundation of China with grant numbers (U21A20485, 62088102), and NSFC General Program with grant number (62176215).}
}
\address{
$^{1}$Institute of Artificial Intelligence and Robotics, Xi'an Jiaotong University, Xi'an, China \\
$^{2}$School of Electrical Engineering, Xi’an University of Technology, Xi'an, China \\
$^{3}$Westlake University Institute of Advanced Technology, Westlake Institute for Advanced Study
}

\begin{document}
\ninept

\maketitle
%


\begin{abstract}
Cross-domain few-shot classification (CDFSC) is a challenging and tough task due to the significant distribution discrepancies across different domains. 
To address this challenge, many approaches aim to learn transferable representations.
Multilayer perceptron (MLP) has shown its capability to learn transferable representations in various downstream tasks, such as unsupervised image classification and supervised concept generalization. 
However, its potential in the few-shot settings has yet to be comprehensively explored.
In this study, we investigate the potential of MLP to assist in addressing the challenges of CDFSC. 
Specifically, we introduce three distinct frameworks incorporating MLP in accordance
with three types of few-shot classification methods to verify the effectiveness of MLP.
We reveal that MLP can significantly enhance discriminative capabilities and alleviate distribution shifts, which can be supported by our expensive experiments involving 10 baseline models and 12 benchmark datasets.
Furthermore, our method even compares favorably against other state-of-the-art CDFSC algorithms.

\end{abstract}

\begin{keywords}
Multilayer Perceptron, cross-domain few-shot classification, distribution shift
\end{keywords}
%

\section{Introduction}
\label{sec:intro}

The introduction of MLP projector after the encoder, initially pioneered in SimCLR~\cite{chen2020simple}, is claimed to effectively mitigate information loss caused by contrastive loss, and has become a widely adopted technique in recent advancements of unsupervised learning frameworks, aimed at enhancing the discriminative capability~\cite{chen2020simple, chen2020improved}.
Similarly, in the context of supervised learning, recent works have explored the integration of contrastive loss and MLP projectors to enhance transferability and improve model performance ~\cite{islam2021broad, khosla2020supervised}.
However, Wang et al.~\cite{wang2022revisiting} argue that previous works overlooked the ablation of the MLP and incorrectly attributed the enhanced transfer performance solely to the contrastive mechanism within the loss function.
Through the utilization of the concept generalization task, they determined that the inclusion of the MLP maintains the intra-class variation present in the pre-training dataset. This indicates the preservation of more instance discriminative information, ultimately resulting in beneficial effects for transfer learning.

\begin{figure}[htbp]
  \centering
  \includegraphics[width=0.472 \textwidth]{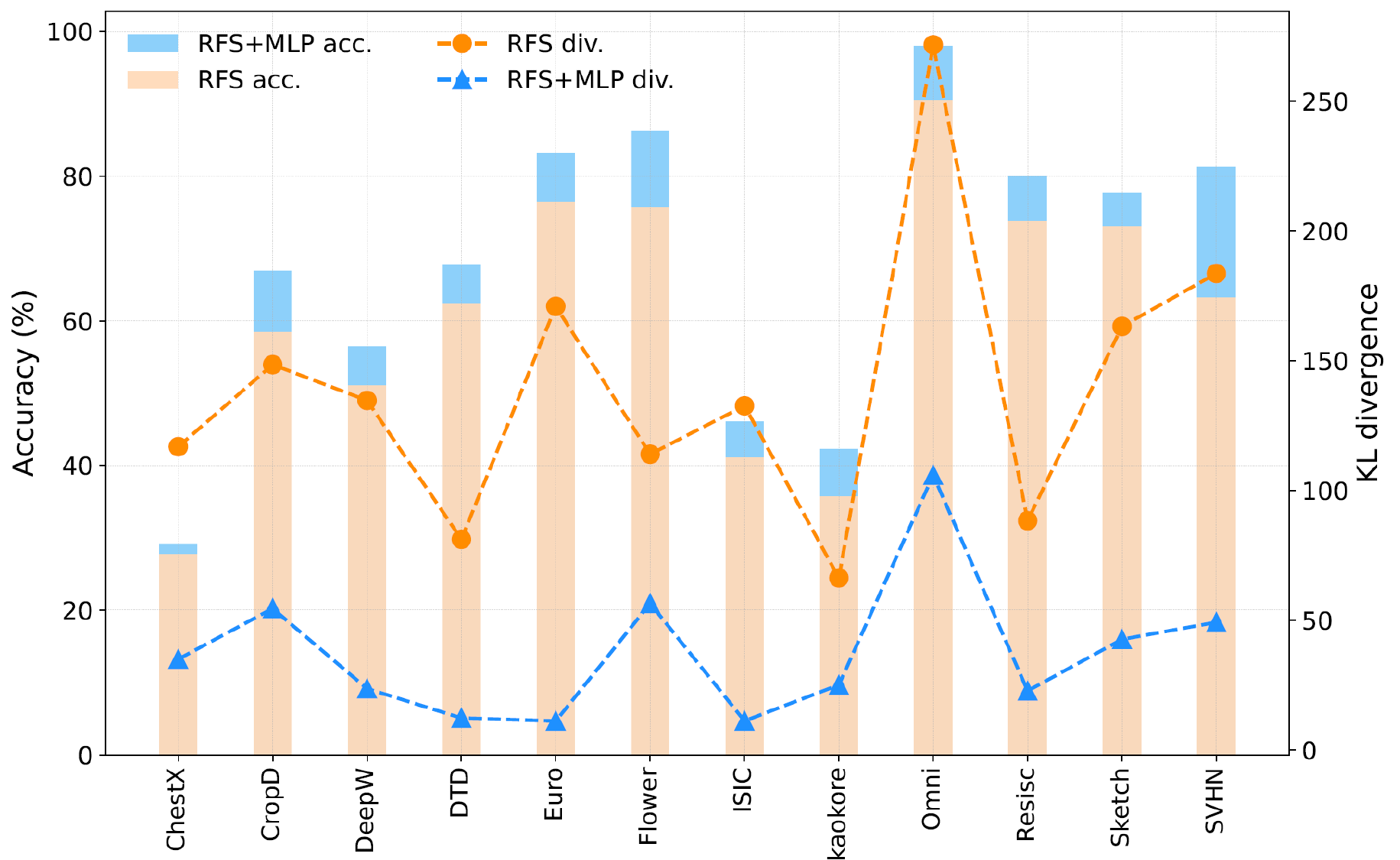}
  \caption{Comparisons of RFS and RFS with MLP on 12 testing datasets. KL divergence (div.) and accuracy (acc.) are reported. KL divergence quantifies the distribution divergence between the pre-training dataset~(\textit{i.e.,} miniImageNet) and each testing dataset.}
  \label{fig:div}
\end{figure}

Although MLP has demonstrated its capability to learn transferable representations in various downstream tasks, the effectiveness of MLP in the few-shot setting has not been explored well.
To validate the transferability of MLP in the few-shot setting, we extend MLP to the task of CDFSC.  
In CDFSC, the goal is to learn from base classes sampled from the training distribution and generalize to novel classes sampled from a distinct testing distribution.
To demonstrate MLP's effectiveness in mitigating distribution shift, we conduct an empirical experiment.
Specifically, we adopt a classical few-shot classification algorithm RFS~\cite{tian2020rethinking} as the baseline, and incorporate MLP before the classifier during the pre-training phase.
As depicted in Figure~\ref{fig:div}, MLP shows the potential to reduce the distribution discrepancy between the pre-training and evaluation datasets, indicating its effectiveness in aligning the feature distributions.

Motivated by this finding, we further introduce three distinct frameworks incorporating MLP in accordance with three types of few-shot classification (FSC) methods.
Specifically, we incorporate an MLP projector before the classifier during the pre-training phase. During the testing phase, we remove the MLP projector and evaluate the transferability of the feature extractor.  
Experimental results on MLP show three interesting findings: 1) MLP helps obtain better discriminative ability. 2) The batch normalization layer in MLP plays the most crucial role in improving transferability. 3) MLP is more efficient when the backbone is more sophisticated. Furthermore, experimental results confirm that adding an MLP into the FSC methods can consistently improve the transferability of the model in the cross-domain setting.
Our main contributions include:

\begin{figure*}[htbp]
  \centering
  \includegraphics[width=\textwidth]{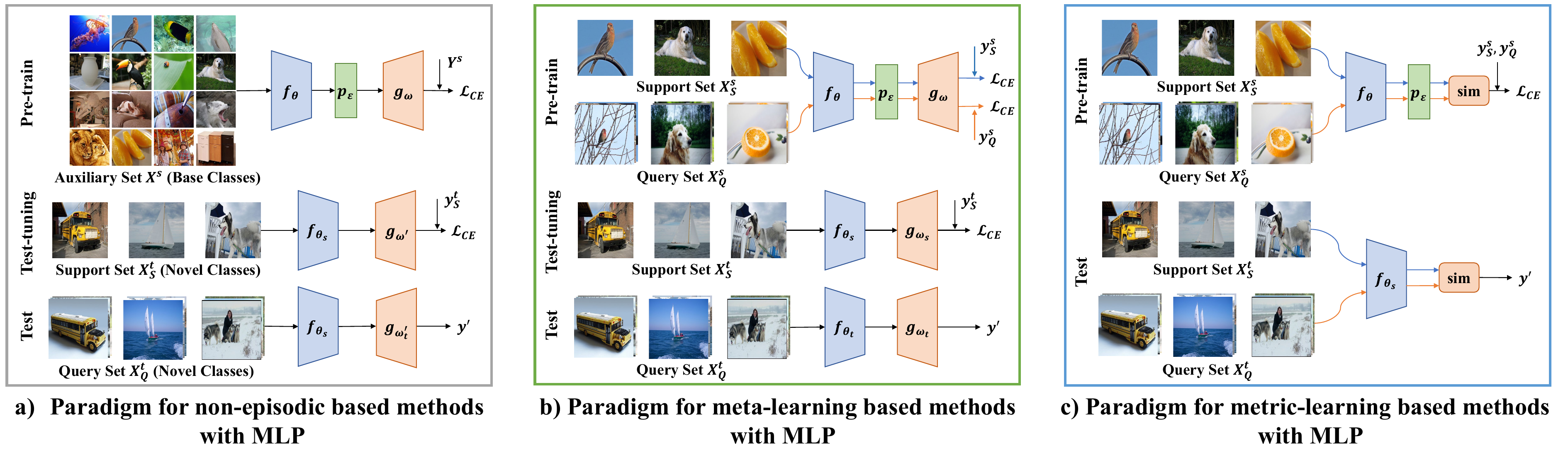}
  \caption{The frameworks with MLP in accordance with three types of few-shot classification methods. Firstly, we pre-train a feature extractor $f$, an MLP projector $p$ and a classifier $g$. $sim$ denotes the classifier that computes similarity, \textit{e.g.}, the nearest-neighbor classifier. MLP projector is removed in the meta-testing phase for evaluating the performance of the pre-trained feature extractor $f$.}
  \label{fig:model}
\end{figure*}

\begin{itemize}
\item We initiate the first known and comprehensive effort to study MLP in CDFSC, and further introduce three distinct frameworks in accordance with three types of few-shot classification methods to verify the effectiveness of MLP.

\item We empirically demonstrate that MLP helps existing few-shot classification algorithms significantly improve cross-domain generalization performance on 12 datasets and even compare favorably against state-of-the-art CDFSC algorithms.

\item Our analyses indicate that MLP helps obtain better discriminative ability and mitigate the distribution shift. Additionally, we find that batch normalization plays the most crucial role in improving transferability.

\end{itemize}

\section{Related Work}
\label{sec:rw}

\subsection{Few-Shot Classification}
Few-Shot Classification (FSC) methods can be divided into three types.
First, meta-learning based methods~\cite{finn2017model, raghu2019rapid} adopt a learning-to-learn paradigm, allowing models to learn from training data and apply this prior learning to improve performance on new tasks or domains.
Second, metric-learning based methods~\cite{vinyals2016matching, snell2017prototypical} employ a learning-to-compare paradigm, allowing models to compare the similarity or the distance of the representations between support set data and query set data.
Finally, non-episodic based methods~\cite{tian2020rethinking, chen2019closer} abandon the learning-to-learn paradigm and adopt the pre-training paradigm from transfer learning. 
In this work, we introduce three distinct frameworks in accordance with these three types of FSC methods by incorporating MLP in cross-domain few-shot setting.

\subsection{Cross-Domain Few-Shot Classification}

One line of CDFSC methods learns various features via data augmentation, such as adversarial task augmentation~\cite{wang2021cross} which learns worst-case tasks, and random crop augmentation~\cite{zhou2023revisiting} which is used for self-supervision learning.
Another family of CDFSC methods aims to align features by using affine transforms~\cite{tseng2020cross} and hyperbolic tangent transformation~\cite{li2022ranking} to minimize the distribution discrepancy between training and test datasets. 
However, these methods often result in complex optimization procedures and expensive computational complexity. 
In this study, we aim to shed new light on improving transferability with MLP, which only introduces a little computational complexity while ensuring simplicity and effectiveness in handling distribution-shift few-shot learning data.

\section{Method}
\label{sec:method}

\begin{table*}[ht]
\centering
\resizebox{\textwidth}{!}{
\begin{tabular}{cccccccccccccccc}  
\toprule
Type & Method  &  {ChestX}  &  {CropD}  &  {DeepW}  &  {DTD}  &  {Euro}  &  {Flower}  &  {ISIC}  &  {Kaokore}  &  {Omni}  & {Resisc}  &  {Sketch}  & {SVHN}  &  Average   \\
\midrule
\multirow{6}{*}{Non-episodic}  
    & BL~\cite{chen2019closer} & 26.98 & 65.44 & 55.27 & 59.58 & 83.81 & 77.59 & 42.69 & 37.24 & 96.24 & 71.34 & 72.60 & 77.49 & 63.86 \\
    & BL+MLP & 28.62 & \underline{70.48} & 57.54 & 61.96 & 84.26 & 81.49 & 42.55 & 40.06 & 98.12 & 75.62 & 75.15 & \underline{84.19} & \textbf{66.67}\smallgreen{+2.81} \\
    & BL++~\cite{chen2019closer} & 25.49 & 48.22 & 50.86 & 51.76 & 75.79 & 66.15 & 40.73 & 32.64 & 87.87 & 63.16 & 63.72 & 64.01 & 55.87 \\
    & BL+++MLP & 28.09 & \underline{65.62} & \underline{57.65} & \underline{64.33} & \underline{85.83} & \underline{81.47} & 44.32 & \underline{38.14} & \underline{97.47} & \underline{74.76} & \underline{75.85} & \underline{82.30} & \textbf{66.32}\smallgreen{+10.45} \\
    & RFS~\cite{tian2020rethinking} & 27.68 & 58.38 & 51.02 & 62.31 & 76.33 & 75.62 & 41.05 & 35.70 & 90.55 & 73.82 & 73.05 & 63.22 & 60.73 \\
    & RFS+MLP & 29.09 & \underline{66.87} & \underline{56.42} & \underline{67.78} & \underline{83.14} & \underline{86.24} & \underline{46.02} & \underline{42.31} & \underline{97.95} & \underline{80.04} & 77.64 & \underline{81.27} & \textbf{67.90}\smallgreen{+7.17} \\
\midrule
\multirow{4}{*}{Meta-learning}      
    & ANIL~\cite{raghu2019rapid} & 24.41 & 48.69 & 46.93 & 46.55 & 63.96 & 61.27 & 37.57 & 31.50 & 84.53 & 58.92 & 61.90 & 58.29 & 52.04 \\
    & ANIL+MLP & 25.02 & \underline{58.10} & 49.35 & \underline{53.30} & \underline{75.73} & \underline{66.45} & 39.19 & 32.61 & 83.02 & \underline{64.62} & 61.70 & 51.97 & \textbf{55.09}\smallgreen{+3.05} \\
    & MTL~\cite{sun2019meta} & 24.15 & 33.27 & 43.14 & 49.43 & 54.27 & 58.18 & 35.56 & 31.36 & 72.77 & 57.93 & 60.53 & 52.72 & 47.78 \\
    & MTL+MLP & 25.19 & \underline{51.23} & 46.06 & 49.41 & \underline{65.19} & 51.34 & 34.74 & 31.59 & \underline{78.84} & 53.97 & 51.62 & 52.01 & \textbf{49.27}\smallgreen{+1.49} \\
\midrule
\multirow{6}{*}{Metric-learning}
    & PN~\cite{snell2017prototypical} & 26.38 & 55.59 & 47.50 & 50.55 & 70.96 & 63.56 & 32.95 & 33.58 & 92.68 & 58.70 & 54.53 & 64.38 & 54.28 \\
    & PN+MLP & 27.07 & \underline{60.76} & 47.83 & 50.49 & 73.22 & 62.63 & 33.76 & 32.00 & 87.06 & 59.02 & 51.85 & 66.69 & \textbf{54.37}\smallgreen{+0.09} \\
    & DN4~\cite{li2019revisiting} & 27.34 & 53.62 & 50.94 & 58.67 & 76.52 & 75.01 & 42.68 & 37.67 & 97.43 & 70.30 & 72.81 & 84.76 & 62.31 \\
    & DN4+MLP & 28.38 & \underline{67.48} & \underline{56.40} & 62.15 & \underline{82.61} & \underline{80.78} & 39.41 & 40.15 & 98.17 & \underline{75.78} & 75.77 & 88.18 & \textbf{66.27}\smallgreen{+3.96} \\
    & CAN~\cite{hou2019cross} & 27.46 & 52.82 & 53.14 & 56.85 & 73.77 & 71.40 & 42.32 & 36.55 & 83.03 & 69.96 & 66.03 & 58.32 & 57.64 \\
    & CAN+MLP & 28.57 & \underline{67.63} & \underline{58.93} & \underline{64.31} & \underline{83.66} & \underline{79.81} & 42.73 & 38.00 & \underline{97.08} & \underline{75.68} & \underline{73.64} & \underline{71.54} & \textbf{65.13}\smallgreen{+7.49} \\
\bottomrule
\end{tabular}
}
\caption{Comparisons with the vanilla baselines in 5-way 5-shot setting on 12 datasets with ResNet12 as the backbone. \textbf{Bold} denotes the best average scores, and \underline{underline} denotes our method outperforms the vanilla methods by a large margin of more than 5.00\%.}
\label{tab:res12_55}
\end{table*}

As shown in Figure \ref{fig:model}, we adapt MLP to three types of FSC methods, \textit{i.e.}, non-episodic, meta-learning and metric-learning based methods. 
Specifically, our method mainly consists of a feature extractor $f(\cdot)$, an MLP projector $p(\cdot)$, and a classifier $g(\cdot)$ or similarity function $sim(\cdot)$. 
The MLP projector consists of two fully connected layers $fc$, a batch normalization layer $BN$, and a ReLU layer $ReLU$, which can be denoted as:

\begin{equation}\begin{aligned}\label{func:step1}
p(\cdot)=fc_2(ReLU(BN(fc_1(\cdot))).
\end{aligned}\end{equation}

In detail, the dimensions of the fully connected layers are set to be the same as the dimensions of the output of the feature extractor.
We introduce three paradigms as follows.

\noindent \textbf{Paradigm for non-episodic based methods with MLP.}
During the training phase, we divide the auxiliary set $X^s$ from the source dataset $D_s$ into batches of samples $T^s=\{X^s_i\}_{i=1}$, and denote the corresponding global labels as $Y^s_i$. 
We use the cross-entropy loss function to train our model:

\begin{equation}\begin{aligned}\label{func:non_pre}
\theta_s, \varepsilon_s, \omega_s=\underset{\theta, \varepsilon, \omega}{\arg \min } \sum_{i=1}^N \mathcal{L}^{\mathrm{CE}}\left(g_\omega\left(p_\varepsilon\left(f_\theta\left(X^s_i\right)\right)\right), Y^s_i\right),
\end{aligned}\end{equation}
where $\theta, \varepsilon, \omega$ are the parameters of the feature extractor, the MLP and the classifier, respectively. During the test-tuning phase, we divide the target dataset into amounts of tasks, and each task can be divided into the a support set $X^t_{S}$ and a query set $X^t_{Q}$.
The support set data are used to fine-tune a new classifier $g_{\omega^{\prime}}$. The optimization function of one loop can be formulated as:

\begin{table}[ht]
\centering
\resizebox{0.42 \textwidth}{!}{
\begin{tabularx}{0.472 \textwidth}{c>{\centering\arraybackslash}X>{\centering\arraybackslash}X>{\centering\arraybackslash}X>{\centering\arraybackslash}X}  
\toprule
Method & RFS & RFS+MLP & RFS & RFS+MLP \\
\midrule
Dataset & \multicolumn{2}{c}{ChestX} & \multicolumn{2}{c}{CropD} \\
\midrule
$D_1$ ($\downarrow$) & 6.85 & \textbf{6.78} & 6.37 & \textbf{5.83} \\
$V$ ($\downarrow$) & 2.73 & \textbf{2.69} & 2.64 & \textbf{2.61} \\
$r$ ($\downarrow$) & 1.21 & \textbf{1.20} & 0.85 & \textbf{0.80} \\ 	
\midrule
Dataset & \multicolumn{2}{c}{Euro} & \multicolumn{2}{c}{ISIC} \\
\midrule
$D_1$ ($\downarrow$) & 3.83 & \textbf{3.57} & 6.09 & \textbf{5.66} \\
$V$ ($\downarrow$) & 1.73 & \textbf{1.51} & 2.35 & \textbf{2.05} \\
$r$ ($\downarrow$) & 0.49 & \textbf{0.43} & 1.04 & \textbf{0.92} \\
\bottomrule
\end{tabularx}
}
\caption{Quantitative analysis of features learned by RFS and RFS+MLP on CD-FSL benchmark. 
$D_1$ is the averaged $L_2$ inner-class distance, and $V$ is the inner-class variance, and $r$ is the ratio between the average of inner-class distance and the inter-class distance. 
\textbf{Bold} denotes the best scores.}
\label{tab:dis}
\end{table}

\begin{equation}\begin{aligned}\label{func:non_tune}
\omega^{\prime}_t 
&= \underset{\omega}{\arg \min } \mathcal{L}^{\mathrm{CE}}\left(g_{{\omega^{\prime}}} \left(f_{\theta_s} \left (X^t_{S} \right) \right), Y^t_{S} \right).
\end{aligned}\end{equation}

After the model is adapted to the support set, the query set data is used to evaluate the model performance.

\noindent \textbf{Paradigm for Meta-learning Based Methods with MLP.}
During the training phase, we divide the auxiliary set $X^s$ from the source dataset $D_s$ into amounts of tasks $T^s=\{ X^s_i \}_{i=1}$, which can be split into support set $T^s_{Si}= \{ {X^s_{Si}} \}_{i=1}$ and query set $T^s_{Qi}=\{ {X^s_{Qi}} \}_{i=1}$. The corresponding local labels are denoted as $y^s_{Si}$ and $y^s_{Qi}$. 
The support set data are used to train our model, thus the optimization function of one inner loop can be formulated as:

\begin{equation}\begin{aligned}\label{func:meta_pre1}
\theta^{\prime}_s, \varepsilon^{\prime}_s, \omega^{\prime}_s=\underset{\theta, \varepsilon, \omega}{\arg \min } \mathcal{L}^{\mathrm{CE}}\left(g_\omega\left(p_\varepsilon\left(f_\theta\left(X^s_{Si}\right)\right)\right), y^s_{Si}\right).
\end{aligned}\end{equation}

The query set data are used to truly update the model, and the optimization function of the outer loop can be formulated as:

\begin{equation}\begin{aligned}\label{func:meta_pre2}
\theta_s, \varepsilon_s, \omega_s=\underset{\theta, \varepsilon, \omega}{\arg \min } \mathcal{L}^{\mathrm{CE}}\left(g_{\omega^{\prime}_s}\left(p_{\varepsilon^{\prime}_s}\left(f_{\theta^{\prime}_s}\left(X^s_{Qi}\right)\right)\right), y^s_{Qi}\right).
\end{aligned}\end{equation}

During the test-tuning and test phase, it also adopts the paradigm of tuning on the support set of the target dataset (like Equation \ref{func:meta_pre1}) and finally testing on the query set of the target dataset.

\begin{figure}[htbp]
  \centering
  \includegraphics[width=0.45 \textwidth]{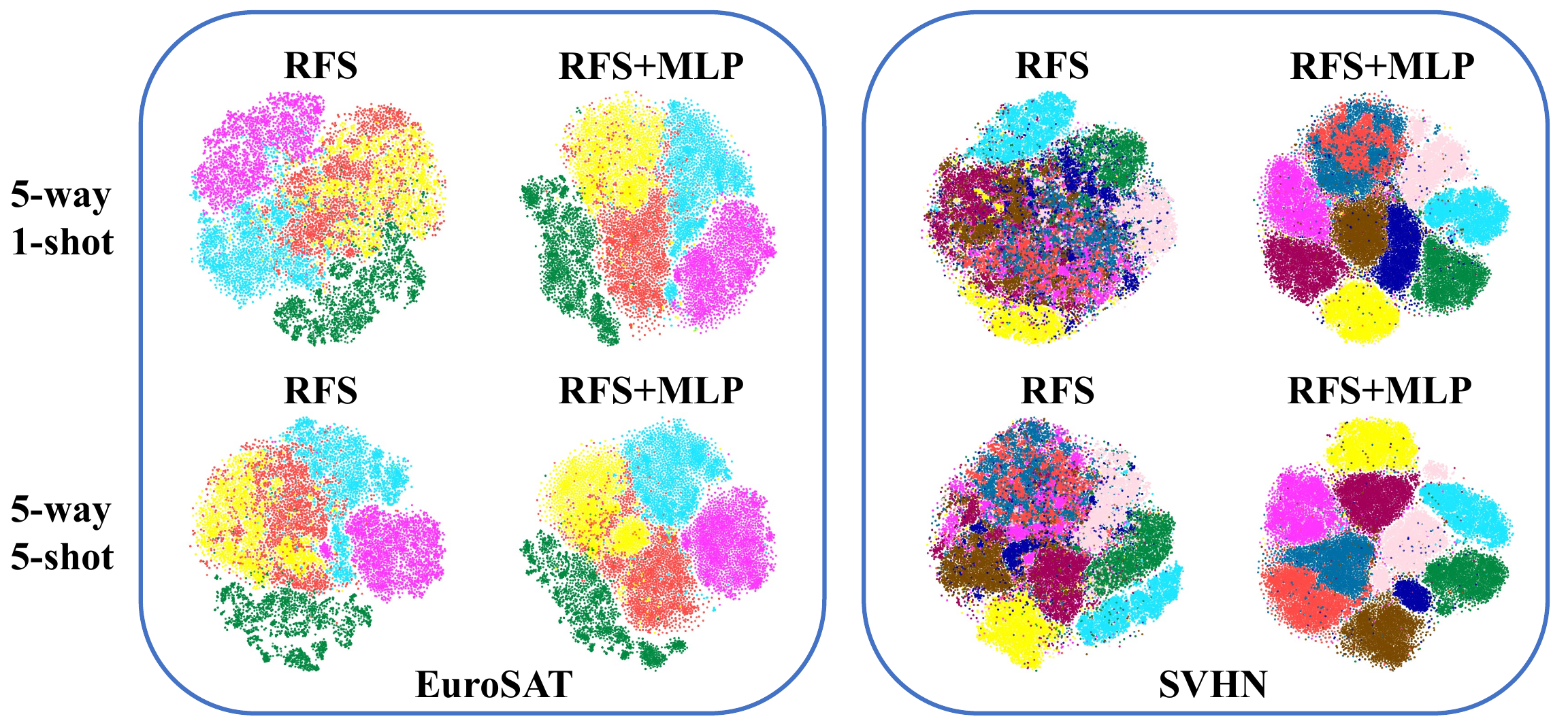}
  \caption{The t-SNE visualization of RFS and RFS+MLP on EuroSAT and SVHN datasets with ResNet12 backbone in 5-way 1-shot and 5-way 5-shot settings. Different colors denote different classes.}
  \label{fig:t-SNE}
\end{figure}

\noindent \textbf{Paradigm for Metric-learning Based Methods with MLP.}
Taking ProtoNet~\cite{snell2017prototypical} as an example, it leverages the support set to calculate the mean centroid of the features.
We denote $X^{s}_{i,j}$ as j-th sample in i-th class from the support set, thus the prototype can be defined as $c^s_i=1/K\sum^K_{j=1} f_{\theta_s}(X^s_{i,j})$, where $K$ denotes the number of images per class. 
Therefore, the probability of a query image $X^s_Q$ of the source dataset belonging to the i-th class can be formulated as:

\begin{equation}\begin{aligned}\label{func:metric_p}
P(y^s_Q=i \mid X^s_Q)
=\frac{\exp \left(-sim\left(p_\varepsilon(f_{\theta}(X^s_Q)), 
c_i\right ) \right)}
{\sum^C_j \exp \left(-sim\left(p_\varepsilon(f_{\theta}(X^s_Q)), c_j\right ) \right)},
\end{aligned}\end{equation}
where $sim$ is a distance function or similarity function. Specifically, during the training phase, the cross-entropy loss is employed to train the model. Also, during the test phase, the nearest-neighbor classifier (like Equation \ref{func:metric_p}) can be conveniently used for prediction.

\section{Experiments}
\label{sec:exp}

\subsection{Experimental Setting}

\noindent \textbf{Datasets.} During the training phase, we train our model on miniImageNet dataset~\cite{vinyals2016matching} from scratch, which consists of 50,000 training images and 10,000 testing images, evenly distributed across 100 classes. We use the train set of miniImageNet for training. During the test-tuning phase, we evaluate the generalization performance on 12 datasets, \textit{i.e.}, CD-FSL benchmark~\cite{guo2020broader} (including ChestX-ray (ChestX)~\cite{wang2017chestx}, CropDisease~(CropD)~\cite{mohanty2016using}, EuroSAT~(Euro)~\cite{helber2019eurosat}, ISIC~\cite{codella2019skin}), DeepWeeds~(DeepW)~\cite{olsen2019deepweeds}, DTD~\cite{cimpoi2014describing}, Flower102 (Flower) \cite{nilsback2008automated}, Kaokore~\cite{tian2020kaokore}, Omniglot~(Omni)~\cite{lake2015human}, Resisc45~(Resisc)~\cite{cheng2017remote}, Sketch~\cite{wang2019learning} and SVHN~\cite{netzer2011reading}, following the setup in~\cite{wang2022revisiting}.

\begin{table}[ht]
\centering
\resizebox{0.45\textwidth}{!}{
\begin{tabular}{cccccc}
\toprule
Method & ChestX & CropD & Euro & ISIC & Average \\
\midrule
MatchingNet~\cite{vinyals2016matching} & 22.40 & 66.39 & 64.45 & 36.74 & 47.50 \\
PN~\cite{snell2017prototypical} & 23.41 & 75.17 & 68.95 & 37.04 & 51.14 \\
PN+FWT~\cite{tseng2020cross} & 23.77 & 85.82 & 67.34 & 38.87 & 53.95 \\
RN~\cite{sung2018learning} & 22.96 & 68.99 & 61.31 & 39.41 & 48.17 \\
RN+FWT~\cite{tseng2020cross} & 23.95 & 75.78 & 69.13 & 38.68 & 51.89 \\
RN+ATA~\cite{wang2021cross} & 24.43 & 78.20 & 71.02 & 40.38 & 53.51 \\
TPN+ATA~\cite{wang2021cross} & 23.60 & 88.15 & 79.47 & 45.83 & 59.26 \\
RDC~\cite{li2022ranking} & 25.91 & 88.90 & 77.15 & 41.28 & 58.31 \\
LDP-net~\cite{zhou2023revisiting} & \textbf{26.67} & \underline{89.40} & \textbf{82.01} & \textbf{48.06} & \textbf{61.54} \\
\midrule
\textbf{BL+MLP (Ours)} & 25.70 & 89.36 & \underline{79.62} & 44.61 & 59.82 \\
\textbf{RFS+MLP (Ours)} & \underline{26.00} & \textbf{89.68} & 78.13 & \underline{46.33} & \underline{60.04} \\
\bottomrule
\end{tabular}
}
\caption{Comparisons with SOTA methods in 5-way 5-shot setting on CD-FSL with ResNet10 as the backbone. \textbf{Bold} denotes the best scores and \underline{underline} denotes the second-best scores.}
\label{tab:sota}
\end{table}

\noindent \textbf{Baselines.} We choose 10 baselines for evaluating the effectiveness of our method. 
For non-episodic based methods, we adopt Baseline~(BL)~\cite{chen2019closer}, Baseline++~(BL++)~\cite{chen2019closer} and RFS~\cite{tian2020rethinking}. 
For meta-learning based methods, we adopt MAML~\cite{finn2017model}, ANIL~\cite{raghu2019rapid}, BOIL~\cite{oh2020boil} and MTL~\cite{sun2019meta}.
For metric-learning based methods, we adopt ProtoNet~(PN)~\cite{snell2017prototypical}, DN4~\cite{li2019revisiting} and CAN~\cite{hou2019cross}.
Furthermore, we compare our method with state-of-the-art~(SOTA) methods, including MatchingNet~\cite{vinyals2016matching}, ProtoNet~(PN)~\cite{snell2017prototypical}, RelationNet~(RN)~\cite{sung2018learning}, FWT~\cite{tseng2020cross}, ATA~\cite{wang2021cross}, RDC~\cite{li2022ranking} and LDP-net~\cite{zhou2023revisiting}.

\noindent \textbf{Experimental Setup.} We mainly use three backbones as the feature extractor to evaluate the effectiveness of MLP, \textit{i.e.}, Conv64F, ResNet12 and ResNet18.
For the training phase, we train all the models for 100 epochs on miniImageNet with SGD optimizer. The learning rate is set to vary within the range of 0.001 to 0.1 across different methods. The momentum rate is 0.9 and the weight decay is 5e-4. The learning rate decay strategy is based on cosine decay. For non-episodic based methods, we set the batch size to 128. For the other two types of methods, we set 2000 episodes in one epoch.
For evaluation, we use the same 600 randomly sampled few-shot episodes for consistency, and report the averaged top-1 accuracy.

\subsection{Analysis Experiments}

The reasons behind the improved transferability achieved by MLP are: 
1) \textbf{MLP can mitigate the distribution shift between the pre-training and evaluation datasets.} 
As shown in Figure~\ref{fig:div}, MLP significantly reduces the KL divergence between the pre-training and evaluation datasets, resulting in higher accuracy.
2) \textbf{MLP can help the model obtain better discriminative ability in terms of cluster compactness.} 
As shown in Table \ref{tab:dis}, the quantitative experiment demonstrates that when employing MLP, there are improvements in lower intra-class distance, intra-class variance, and the ratio between the average intra-class distance and inter-class distance. This indicates that MLP can make the learned representations more discriminable. 
Qualitatively, in Figure \ref{fig:t-SNE}, we can also observe that MLP can help the model obtain better discriminative ability.

\subsection{Comparisons}

\noindent \textbf{Comparisons with vanilla FSC methods.} Table \ref{tab:res12_55} demonstrates that all methods with MLP outperform the vanilla methods. For instance, BL++, RFS and CAN with MLP outperform the vanilla methods by a large margin of 10.45\%, 7.17\% and 7.49\% on 12 datasets for the averaged top-1 accuracy, respectively. 
For some methods, there can be a large improvement on some datasets, \textit{e.g.,} +14.05\% for CAN on Omniglot and +18.05\% for RFS on SVHN, indicating MLP can significantly enhance the generalization performance of existing FSC methods.

\begin{table}[ht]
\centering
\resizebox{0.45 \textwidth}{!}{
\begin{tabular}{cccccc}
\toprule
Example & Input FC & BN & ReLU & Output FC & Acc. \\
\midrule
(a) &  &  &  &  & 60.73 \\
(b) & $\checkmark$ &  &  &  & 61.03 (\textcolor[RGB]{0,100,0}{+0.30}) \\
(c) & $\checkmark$ & $\checkmark$ &  &  & 67.66 (\textcolor[RGB]{0,100,0}{+6.93}) \\
(d) & $\checkmark$ &  & $\checkmark$ &  & 61.14 (\textcolor[RGB]{0,100,0}{+0.41}) \\
(e) & $\checkmark$ & $\checkmark$ &  & $\checkmark$ & 67.14 (\textcolor[RGB]{0,100,0}{+6.41}) \\
(f) & $\checkmark$ &  & $\checkmark$ & $\checkmark$ & 62.05 (\textcolor[RGB]{0,100,0}{+1.32}) \\
(g) & $\checkmark$ & $\checkmark$ & $\checkmark$ &  & 67.61 (\textcolor[RGB]{0,100,0}{+6.88}) \\
\midrule
RFS+MLP & $\checkmark$ & $\checkmark$ & $\checkmark$ & $\checkmark$ & \textbf{67.90 (\textcolor[RGB]{0,100,0}{+7.17})} \\
\bottomrule
\end{tabular}
}
\caption{Ablation on different components of MLP. The average results of 12 datasets are reported. Improvements over the baseline of RFS w/o MLP are marked in green.}
\label{tab:abla}
\end{table}

\begin{figure}[htbp]
  \centering
  \includegraphics[width=0.472 \textwidth]{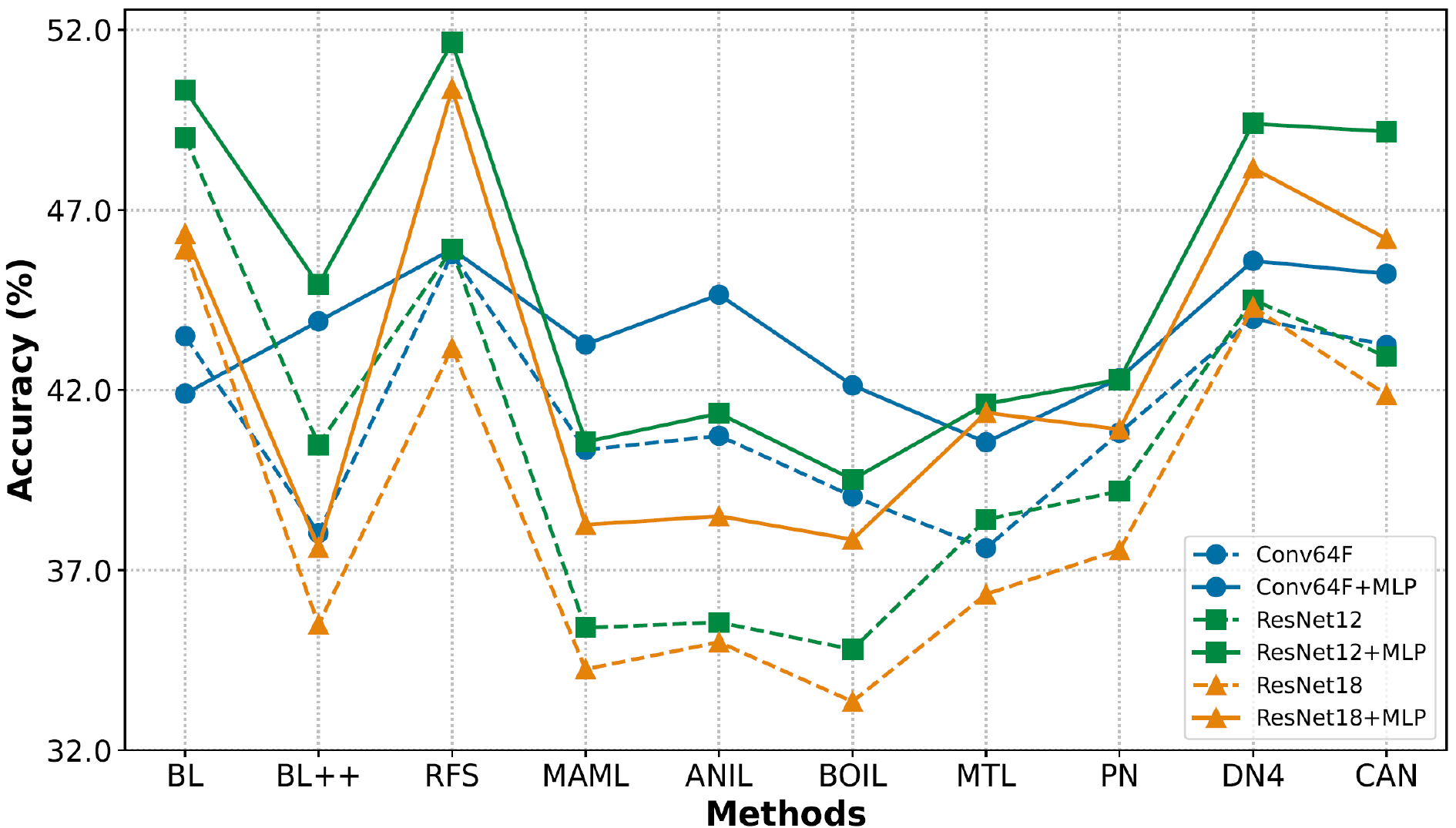}
  \caption{Investigations on various backbones in 5-way 1-shot setting on 12 datasets. Different colors denote different backbones.}
  \label{fig:backbone}
\end{figure}

\noindent \textbf{Comparisons with SOTA CDFSC methods.} 
As shown in Table \ref{tab:sota}, in 5-way 5-shot setting, the averaged result of our method can reach around 60\%, which is superior to and competitive to most of the state-of-the-art CDFSC algorithms. 
Our method incorporates an MLP to learn a more transferable feature extractor. This brings about only a marginal increase in computational complexity, further demonstrating the simplicity and effectiveness of our method.

\subsection{Ablation Study}

\noindent \textbf{Effectiveness of different components in MLP}. As shown in Table \ref{tab:abla}, we observe that integrating all components of MLP achieves the best accuracy among all variants.
Furthermore, we find that Batch Normalization layer plays the most important role in enhancing the transferability of the model, whereas the linear layer and ReLU activation function are comparatively less influential.

\noindent \textbf{Effectiveness of different backbones}. As shown in Figure \ref{fig:backbone}, we observe that MLP is efficient across different backbones, particularly when employed with more sophisticated backbone architectures.
For the Conv64F backbone, most methods with MLP outperform vanilla methods by 0$\sim$3\%, and by 3$\sim$6\% for the ResNet12 and the ResNet18 backbone on 12 datasets.

\section{Conclusion}
\label{sec:conclu}

In this paper, we initiate the first empirical study on using MLP for cross-domain few-shot classification. 
To verify the effectiveness of MLP in the few-shot setting, we introduce three distinct frameworks incorporating MLP in accordance with three types of few-shot classification methods.
Based on empirical results, we identify that the MLP helps the model obtain better discriminative ability and mitigate the distribution shift, which results in good transferability. 
Our findings are confirmed with extensive
experiments on various few-shot classification algorithms and diverse backbone networks.



\bibliographystyle{IEEEbib}
\bibliography{ieee}

\end{document}